%% file: main.tex
\documentclass[10pt,twocolumn,letterpaper]{article}

\usepackage{cvpr}
\usepackage{times}
\usepackage{epsfig}
\usepackage{graphicx}
\usepackage{amsmath}
\usepackage{amssymb}
\usepackage{comment}
\usepackage{caption}
\usepackage{stfloats}
\usepackage{multirow}

\newcommand{\OURS}{SG-NN}
\newcommand{\OURSFULL}{Sparse Generative Neural Network}

\usepackage[pagebackref=true,breaklinks=true,letterpaper=true,colorlinks,bookmarks=false]{hyperref}

\cvprfinalcopy 

\def\cvprPaperID{XXXX} 

\ifcvprfinal\fi
\begin{document}

\title{\OURS{}: Sparse Generative Neural Networks for \\ Self-Supervised Scene Completion of RGB-D Scans}

\author{
	Angela Dai \hspace{2cm} Christian Diller \hspace{2cm} Matthias Nie{\ss}ner \vspace{0.1cm} \\
	Technical University of Munich \\
}

\twocolumn[{%
	\renewcommand\twocolumn[1][]{#1}%
	\maketitle
	\begin{center}
		\vspace{-0.2cm}
		\includegraphics[width=0.94\linewidth]{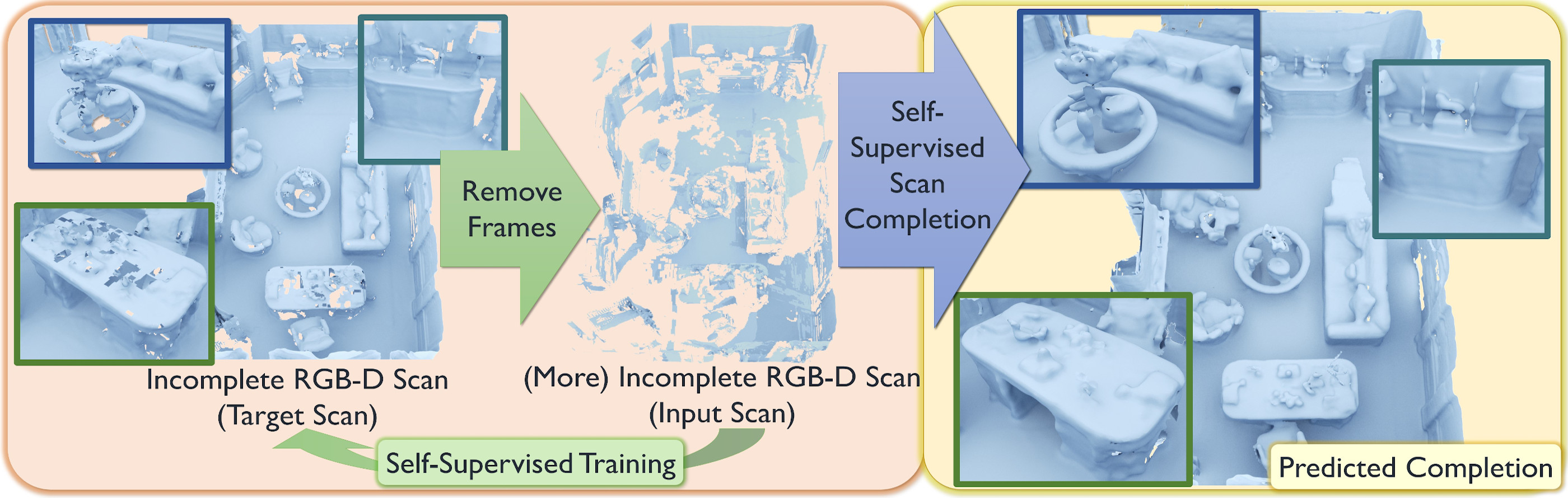}
		\vspace{-0.2cm}
		\captionof{figure}{
		Our method takes as input a partial RGB-D scan and predicts a high-resolution 3D reconstruction while predicting unseen, missing geometry. 
		Key to our approach is its self-supervised formulation, enabling training solely on real-world, incomplete scans. 
		This not only obviates the need for synthetic ground truth, but is also capable of generating more complete scenes than any single target scene seen during training. 
		To achieve high-quality surfaces, we further propose a new sparse generative neural network, capable of generating large-scale scenes at much higher resolution than existing techniques.
		}
		\label{fig:teaser}
	\end{center}
}]

\thispagestyle{empty}

\input{0abstract}

\input{1intro}
\input{2relatedwork}

\input{3method}

\input{4selfsup}

\input{5arch}

\input{6results}

\input{7conclusion}

\section*{Acknowledgments}
This work was supported by the ZD.B, a Google Research Grant, a TUM-IAS Rudolf M{\"o}{\ss}bauer Fellowship, an NVidia Professorship Award, and the ERC Starting Grant Scan2CAD (804724).

{\small
\bibliographystyle{ieee_fullname}
\bibliography{main}
}

\clearpage
\newpage
\begin{appendix}

\input{appendix}
\end{appendix}

\end{document}

%% file: 0abstract.tex
\begin{abstract}
We present a novel approach that converts partial and noisy RGB-D scans into high-quality 3D scene reconstructions by inferring unobserved scene geometry.
Our approach is fully self-supervised and can hence be trained solely on real-world, incomplete scans.
To achieve self-supervision, we remove frames from a given (incomplete) 3D scan in order to make it even more incomplete; self-supervision is then formulated by correlating the two levels of partialness of the same scan while masking out regions that have never been observed.
Through generalization across a large training set, we can then predict 3D scene completion without ever seeing any 3D scan of entirely complete geometry.
Combined with a new 3D sparse generative  neural network architecture, our method is able to predict highly-detailed surfaces in a coarse-to-fine hierarchical fashion, generating 3D scenes at $2$cm resolution, more than twice the resolution of existing state-of-the-art methods as well as outperforming them by a significant margin in reconstruction quality.\footnote{Source code \href{https://github.com/angeladai/sgnn}{available here}.}
\pagebreak
\end{abstract}
\vspace{-1.35cm}

%% file: 1intro.tex
\section{Introduction}
\label{sec:intro}

In recent years, we have seen incredible progress on RGB-D reconstruction  of indoor environments using commodity RGB-D sensors such as the Microsoft Kinect, Google Tango, or Intel RealSense \cite{newcombe2011kinectfusion,izadi2011kinectfusion,niessner2013hashing,whelan2015elasticfusion,choi2015robust,dai2017bundlefusion}.
However, despite remarkable achievements in RGB-D tracking and reconstruction quality, a fundamental challenge still remains -- the incomplete nature of resulting 3D scans caused by inherent occlusions due to the physical limitations of the scanning process; i.e., even in a careful scanning session it is inevitable that some regions of a 3D scene remain unobserved.
This unfortunately renders the resulting reconstructions unsuitable for many applications, not only those that require quality 3D content, such as video games or AR/VR, but also robotics where a completed 3D map significantly facilitates tasks such as grasping or querying 3D objects in a 3D environment.

In order to overcome the incomplete and partial nature of 3D reconstructions, various  geometric inpainting techniques have been proposed, for instance, surface interpolation based on the Poisson equation \cite{kazhdan2006poisson,kazhdan2013screened} or CAD shape-fitting techniques \cite{avetisyan2019scan2cad,avetisyan2019end2end,dahnert2019embedding}.
A very recent direction leverages generative deep neural networks, often focusing volumetric representations for shapes~\cite{dai2017complete} or entire scenes~\cite{song2017ssc,dai2018scancomplete}.
These techniques show great promise since they can learn generalized patterns in a large variety of environments; however, existing data-driven scene completion methods rely on supervised training, requiring fully complete ground truth 3D models, thus depending on large-scale synthetic datasets such as ShapeNet~\cite{shapenet2015} or SUNCG~\cite{song2017ssc}.
As a result, although we have seen impressive results from these approaches on synthetic test sets, domain transfer and application to real-world 3D scans remains a major limitation.

In order to address the shortcomings of supervised learning techniques for scan completion, we propose a new self-supervised completion formulation that can be trained only on (partial) real-world data.
Our main idea is to learn to generate more complete 3D models from less complete data, while masking out any unknown regions; that is, from an existing RGB-D scan, we use the scan as the target and remove frames to obtain a more incomplete input. 
In the loss function, we can now correlate the difference in partialness between the two scans, and constrain the network to predict the delta while masking out unobserved areas.
Although there is no single training sample which contains a fully-complete 3D reconstruction, we show that our network can nonetheless generalize to predict high levels of completeness through a combined aggregation of patterns across the entire training set.
This way, our approach can be trained without requiring any fully-complete ground truth counterparts that would make generalization through a synthetic-to-real domain gap challenging.

Furthermore, we propose a new sparse generative neural network architecture that can predict high-resolution geometry in a fully-convolutional fashion.
For training, we propose a progressively growing network architecture trained in coarse-to-fine fashion; i.e., we first predict the 3D scene at a low resolution, and then continue increasing the surface resolution through the training process.
We show that our self-supervised, sparse generative approach can outperform state-of-the-art fully-supervised methods, despite their access to much larger quantities of synthetic 3D data.

We present the following main contributions:
\begin{itemize}
    \item A self-supervised approach for scene completion, enabling training solely on incomplete, real-world scan data while predicting geometry more complete than any seen during training, by leveraging common patterns in the deltas of incompleteness.
    \item A generative formulation for sparse convolutions to produce a sparse truncated signed distance function representation at high resolution: we formulate this hierarchically to progressively generate a 3D scene in end-to-end fashion
\end{itemize}

%% file: 2relatedwork.tex
\section{Related Work}
\label{sec:relatedWork}

\paragraph{RGB-D Reconstruction}

Scanning and reconstructing 3D surfaces has a long history across several research communities. 
With the increase in availability of commodity range sensors, capturing and reconstructing 3D scenes has become a vital area of research.
One seminal technique is the volumetric fusion approach of Curless and Levoy \cite{curless1996volumetric}, operating on truncated signed distance fields to produce a surface reconstruction. 
It has been adopted by many state-of-the-art real-time reconstruction methods, from  KinectFusion~\cite{newcombe2011kinectfusion,izadi2011kinectfusion} to VoxelHashing~\cite{niessner2013hashing} and  BundleFusion~\cite{dai2017bundlefusion}, as well as state-of-the-art offline reconstruction approaches~\cite{choi2015robust}.

These methods have produced impressive results in tracking and scalability of 3D reconstruction from commodity range sensors.  
However, a significant limitation that still remains is the partial nature of 3D scanning; i.e., a perfect scan is usually not possible due to occlusions and unobserved regions and thus, the resulting 3D representation cannot reach the quality of manually created 3D assets.

\paragraph{Deep Learning on 3D Scans}
With recent advances in deep learning and the improved availability of large-scale 3D scan datasets such as ScanNet~\cite{dai2017scannet} or Matterport~\cite{Matterport3D}, learned approaches on 3D data can be used for a variety of tasks like classification, segmentation, or completion. 

Many current methods make use of convolutional operators that have been shown to work well on 2D data. 
When extended into 3D, they operate on regular grid representations such as distance fields \cite{dai2017complete} or occupancy grids \cite{maturana2015voxnet}.
Since dense volumetric grids can come with high computational and memory costs, several recent approaches have leveraged the sparsity of the 3D data for discriminative 3D tasks.
PointNet~\cite{qi2017pointnet, qi2017pointnetplusplus} introduced a deep network architecture for learning on point cloud data for semantic segmentation and classification tasks.
Octree-based approaches have also been developed \cite{riegler2017OctNet,wang2017cnn,wang2018adaptive} that have been shown to be very memory efficient; however, generative tasks involving large, varying-sized environments seems challenging and octree generation has only been shown for single ShapeNet-style objects \cite{riegler2017octnetfusion,ogn2017}.
Another option leveraging the sparsity of 3D geometric data is through sparse convolutions \cite{graham2017submanifold,graham20183dsemantic,choy20194d}, which have seen success in discriminative tasks such as semantic segmentation, but not in the context of generative 3D modeling tasks, where the overall structure of the scene is unknown.

\paragraph{Shape and Scene Completion}
Completing 3D scans has been well-studied in geometry processing.
Traditional methods, such as Poisson Surface Reconstruction~\cite{kazhdan2006poisson,kazhdan2013screened}, locally optimize for a surface to fit to observed points, and work well for small missing regions.
Recently, various deep learning-based approaches have been developed with greater capacity for learning global structures of shapes, enabling compelling completion of larger missing regions in scans of objects~\cite{wu20153d, dai2017complete, han2017complete,Wang2017ShapeIU,Park2019DeepSDFLC}.
Larger-scale completion of scans has been seen with SSCNet~\cite{song2017ssc}, operating on a depth image of a scene, and ScanComplete~\cite{dai2018scancomplete}, which demonstrated scene completion on room- and building floor-scale scans.
However, both these approaches operate on dense volumetric grids, significantly limiting their output resolutions.
Moreover, these approaches are fully supervised with complete 3D scene data, requiring training on synthetic 3D scene data (where complete ground truth is known), in order to complete real-world scans.

An alternative approach for shape completion could through leveraging a single implicit latent space, as in DeepSDF~\cite{Park2019DeepSDFLC} or Occupancy Networks~\cite{OccupancyNetworks}; however, it still remains a challenge as to how to scale a single latent space  to represent large, varying-sized environments.

%% file: 3method.tex
\begin{figure*}[bp]
	\centering
	\includegraphics[width=0.94\linewidth]{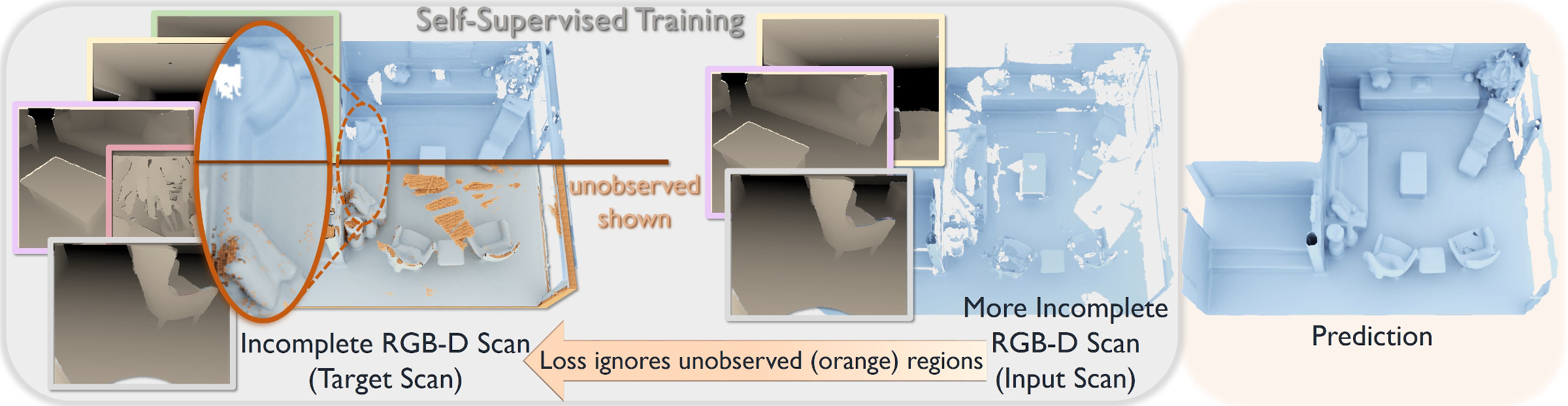}
	\vspace{-0.2cm}
	\caption{
	    Our self-supervision approach for scan completion learns through deltas in partialness of RGB-D scans. 
	    From a given (incomplete) RGB-D scan, on the left, we produce a more incomplete version of the scan by removing some of its depth frames (middle).
	    We can then train to complete the more incomplete scan (middle) using the original scan as a target (left), while masking out unobserved regions in the target scene (in orange).
	    This enables our prediction to produce scenes that are more complete than the target scenes seen during training, as the training process effectively masks out incompleteness.
    }
	\label{fig:selfsup}
\end{figure*}

\section{Method Overview}
\label{sec:overview}

From an RGB-D scan of a 3D scene, our method learns to generate a high-quality reconstruction of the complete 3D scene, in a self-supervised fashion.
The input RGB-D scan is represented as a truncated signed distance field (TSDF), as a sparse set of voxel locations within truncation and their corresponding distance values.
The output complete 3D model of the scene is also generated as a sparse TSDF (similarly, a set of locations and per-voxel distances), from which a mesh can be extracted by Marching Cubes~\cite{lorensen1987marching}.

We design the 3D scene completion as a self-supervised process, enabling training purely on real-world scan data without requiring any fully-complete ground truth scenes. 
Since real-world scans are always incomplete due to occlusions and physical sensor limitations, this is essential for generating high-quality, complete models from real-world scan data.
To achieve self-supervision, our main idea is to formulate the training from incomplete scan data to less incomplete scan data; that is, from an existing RGB-D scan we can remove frames in order to create a more partial observation of the scene.
This enables learning to complete in regions where scan geometry is known while ignoring regions of unobserved space.
Crucially, our generative model can then learn to generate more complete models than seen in a specific sample of the target data.

To obtain an output high-resolution 3D model of a scene, we propose \OURSFULL s (\OURS{}), a generative model to produce a sparse surface representation of a scene.
We build upon sparse convolutions~\cite{graham2017submanifold,graham20183dsemantic,choy20194d}, which have been shown to produce compelling semantic segmentation results on 3D scenes by operating only on surface geometry.
In contrast to these discriminative tasks where the geometric structure is given as input, we develop our \OURS{} to generate new, unseen 3D geometry suitable for generative 3D modeling tasks.
This is designed in coarse-to-fine fashion, with a progressively growing network architecture which predicts each next higher resolution, finally predicting a high-resolution surface as a sparse TSDF.
Since our sparse generative network operates in a fully-convolutional fashion, we can operate on 3D scans of varying spatial sizes.

%% file: 4selfsup.tex
\begin{figure*}[bp]
	\centering
	\includegraphics[width=0.94\linewidth]{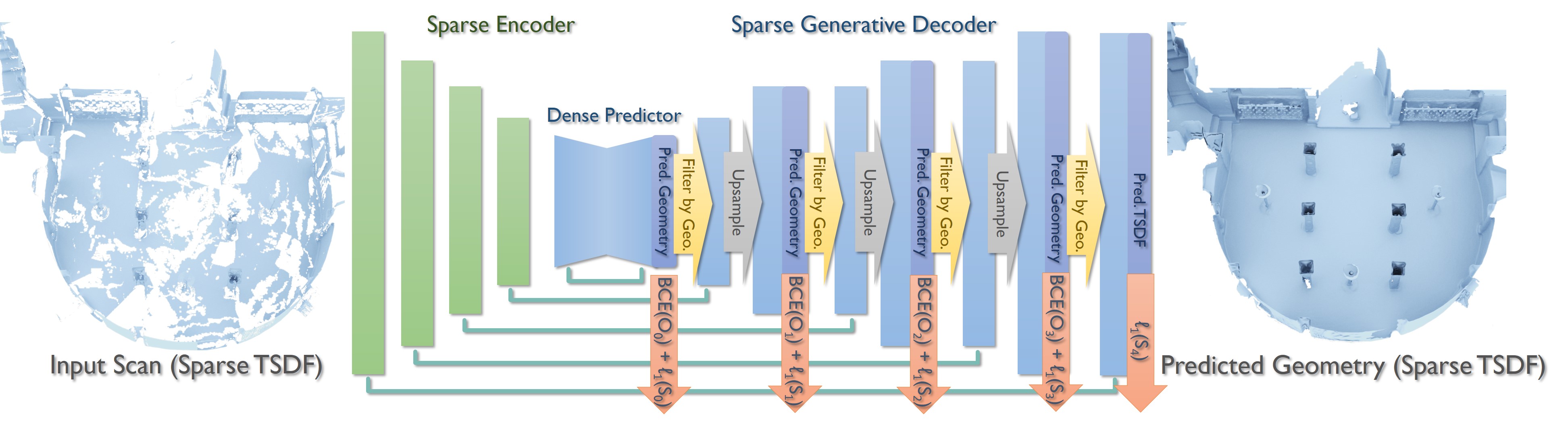}
	\vspace{-0.4cm}
	\caption{
	    Our \OURSFULL{} architecture for the task of scan completion. An input scan is encoded using a series of sparse convolutions, each set reducing the spatial dimensions by a factor of two. 
	    To generate high-resolution scene geometry, the coarse encoding is converted to a dense representation for a coarse prediction of the complete geometry. 
	    The predicted coarse geometry is converted to a sparse representation and input to our sparse, coarse-to-fine hierarchy, where each level of the hierarchy predicts the geometry of the next resolution (losses indicated in orange).
	    The final output is a TSDF represented by sparse set of voxel locations and their corresponding distance values.
    }
	\label{fig:arch}
\end{figure*}

\section{Self-Supervised Completion}
\label{sec:selfsup}

Our approach for self-supervision of scene completion of RGB-D scans is based on learning how to complete scan geometry in regions that have been seen, while ignoring unobserved regions.
To this end, we can generate input and target TSDFs with similar scanning patterns as real-world scans; from an input scan composed of RGB-D frames $\{f_0,...f_n\}$, we can generate the target TSDF $\mathcal{S}_{\mathrm{target}}$ through volumetric fusion~\cite{curless1996volumetric} of $\{f_0,...f_n\}$, and the input TSDF $\mathcal{S}_{\mathrm{input}}$ through volumetric fusion of a subset of the original frames $\{f_k\}\subset \{f_0,...f_n\}$.

This produces input incomplete scans that maintain scanned data characteristics, as well as a correspondence between $\mathcal{S}_{\mathrm{input}}$ and $\mathcal{S}_{\mathrm{target}}$ going from a more incomplete scan to a less incomplete scan.
Since $\mathcal{S}_{\mathrm{target}}$ remains nonetheless incomplete, we do not wish to use all of its data as the complete target for supervision, as this could result in contradictory signals in the training set (e.g., table legs have been seen in one scan but not in another, then it becomes unclear whether to generate table legs).

Thus, to effectively learn to generate a complete 3D model beyond even the completeness of the target training data, we formulate the completion loss only on observed regions in the target scan.
That is, the loss is only considered in regions where $\mathcal{S}_{\mathrm{target}}(v) > -\tau$, for a voxel $v$ with $\tau$ indicating the voxel size.
Figure~\ref{fig:selfsup} shows an example $\mathcal{S}_{\mathrm{input}}$, $\mathcal{S}_{\mathrm{target}}$, and prediction, with this self-supervision setup, we can learn to predict geometry that was unobserved in $\mathcal{S}_{\mathrm{target}}$, e.g., occluded regions behind objects. 

\subsection{Data Generation}
\label{subsec:datagen}

As input we consider an RGB-D scan comprising a set of depth images and their $6$-DoF camera poses.
For real-world scan data we use the Matterport3D~\cite{Matterport3D} dataset, which contains a variety of RGB-D scans taken with a Matterport tripod setup.
Note that for Matterport3D, we train and evaluate on the annotated room regions, whereas the raw RGB-D data is a sequence covering many different rooms, so we perform an approximate frame-to-room association by taking frames whose camera locations lie within the room.

From a given RGB-D scan, we construct the target scan $\mathcal{S}_{\mathrm{target}}$ using volumetric fusion~\cite{curless1996volumetric} with $2$cm voxels and truncation of $3$ voxels.
A subset of the frames is taken by randomly removing $\approx 50\%$ of the frames (see Section~\ref{sec:results} for more analysis of varying degrees of incompleteness in $\mathcal{S}_{\mathrm{input}}$, $\mathcal{S}_{\mathrm{target}}$).
We can then again use volumetric fusion to generate a more incomplete version of the scan \textbf{$\mathcal{S}_{\mathrm{input}}$}.

At train time, we consider cropped views of these scans for efficiency, using random crops of size $64\times 64\times 128$ voxels.
The fully-convolutional nature of our approach enables testing on full scenes of varying sizes at inference time.

%% file: 5arch.tex
\section{Generating a Sparse 3D Scene Representation}
\label{sec:arch}

The geometry of a 3D scene occupies a very sparse set of the total 3D extent of the scene, so we aim to generate a 3D representation of a scene in a similarly sparse fashion.
Thus we propose \OURSFULL s (\OURS{}) to hierarchically generate a sparse, truncated signed distance field representation of a 3D scene, from which we can extract the isosurface as the final output mesh.

An overview of our network architecture for the scene completion task is shown in Figure~\ref{fig:arch}.
The model is designed in encoder-decoder fashion, with an input partial scan first encoded to representative features at low spatial resolution, before generating the final TSDF output.

A partial scan, represented as a TSDF, is encoded with a series of 3D sparse convolutions~\cite{graham2017submanifold,graham20183dsemantic} which operate only on the locations where the TSDF is within truncation distance and using the distance values as input features.
Each set of convolutions spatially compresses the scene by a factor of two. 
Our generative model takes the encoding of the scene and converts the features into a (low-resolution) dense 3D grid.
The dense representation enables prediction of the full scene geometry at very coarse resolution; here, we use a series of dense 3D convolutions to produce a feature map $F_0$ from which we also predict coarse occupancy $O_0$ and TSDF $S_0$ representations of the complete scene. 
We then construct a sparse representation of the predicted scene based on $O_0$: the features input to the next level are composed as $\textrm{concat}(F_k, O_k, S_k)\; \forall\; \textrm{sigmoid}(O_k(v)) > 0.5$.
This can then be processed with sparse convolutions, then upsampled by a factor of two to predict the scene geometry at the next higher resolution.
This enables generative, sparse predictions in a hierarchical fashion.
To generate the final surface geometry, the last hierarchy level of our \OURS{}  outputs sparse $O_n$, $S_n$, and $F_n$, which are then input to a final set of sparse convolutions to refine and predict the output signed distance field values.

\paragraph{Sparse skip connections.} For scene completion, we also leverage skip connections between the encoder and decoder parts of the network architecture, connecting feature maps of same spatial resolution.
This is in the same spirit as U-Net~\cite{ronneberger2015u}, but in our case the encoder and decoder features maps are both sparse and typically do not contain the same set of sparse locations.
Thus we concatenate features from the set of source locations which are shared with the destination locations, and use zero feature vectors for the destination locations which do not exist in the source.

\begin{figure}[!htb] 
	\centering
	\includegraphics[width=\linewidth]{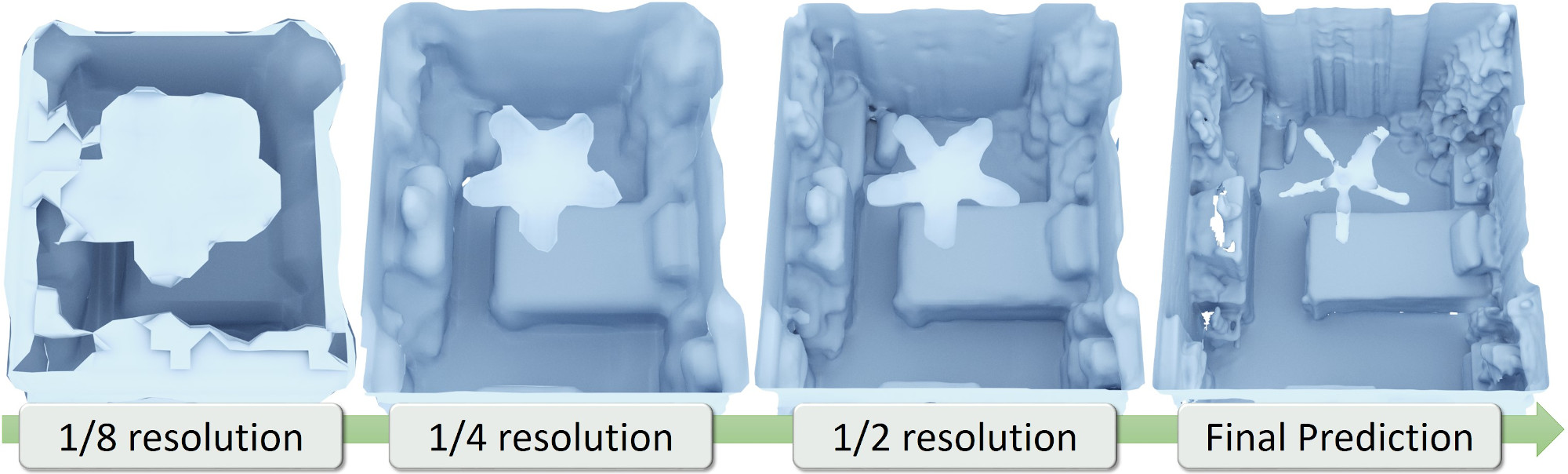}
	\vspace{-0.7cm}
	\caption{
        Progressive generation of a 3D scene using our \OURS{} which formulates a generative model to predict a sparse TSDF as output.
	\vspace{-0.5cm}
    }
	\label{fig:proggen}
\end{figure}

\paragraph{Progressive Generation.}
In order to encourage more efficient and stable training, we train our generative model progressively, starting with the lowest resolution, and introducing each successive hierarchy level after $N_{\textrm{level}}$ iterations.
Each hierarchy level predicts the occupancy and TSDF of the next level, enabling successive refinement from coarse predictions, as shown in Figure~\ref{fig:proggen}.

\paragraph{Loss.} We formulate the loss for the generated scene geometry on the final predicted TSDF locations and values, using an $\ell_1$ loss  with the target TSDF values at those locations.
Following \cite{dai2017complete}, we log-transform the TSDF values of the predictions and the targets before applying the $\ell_1$ loss, in order to encourage more accurate prediction near the surface geometry.
We additionally employ proxy losses at each hierarchy level for outputs $O_k$  and $S_k$, using binary cross entropy with target occupancies and $\ell_1$ with target TSDF values, respectively. 
This helps avoid a trivial solution of zero loss for the final surface with no predicted geometry.
Note that for our self-supervised completion, we compute these losses only in regions of observed target values, as described in Section~\ref{sec:selfsup}.

\subsection{Training}
\label{subsec:training}
We train our \OURS{} on a single NVIDIA GeForce RTX 2080, using the Adam optimizer with a learning rate of $0.001$ and batch size of $8$.
We use $N_{\textrm{level}}=2000$ iterations for progressive introduction of each higher resolution output, and train our model for $\approx 40$ hours until convergence.

%% file: 6results.tex
\begin{figure*}[tp]
	\centering
	\includegraphics[width=0.85\linewidth]{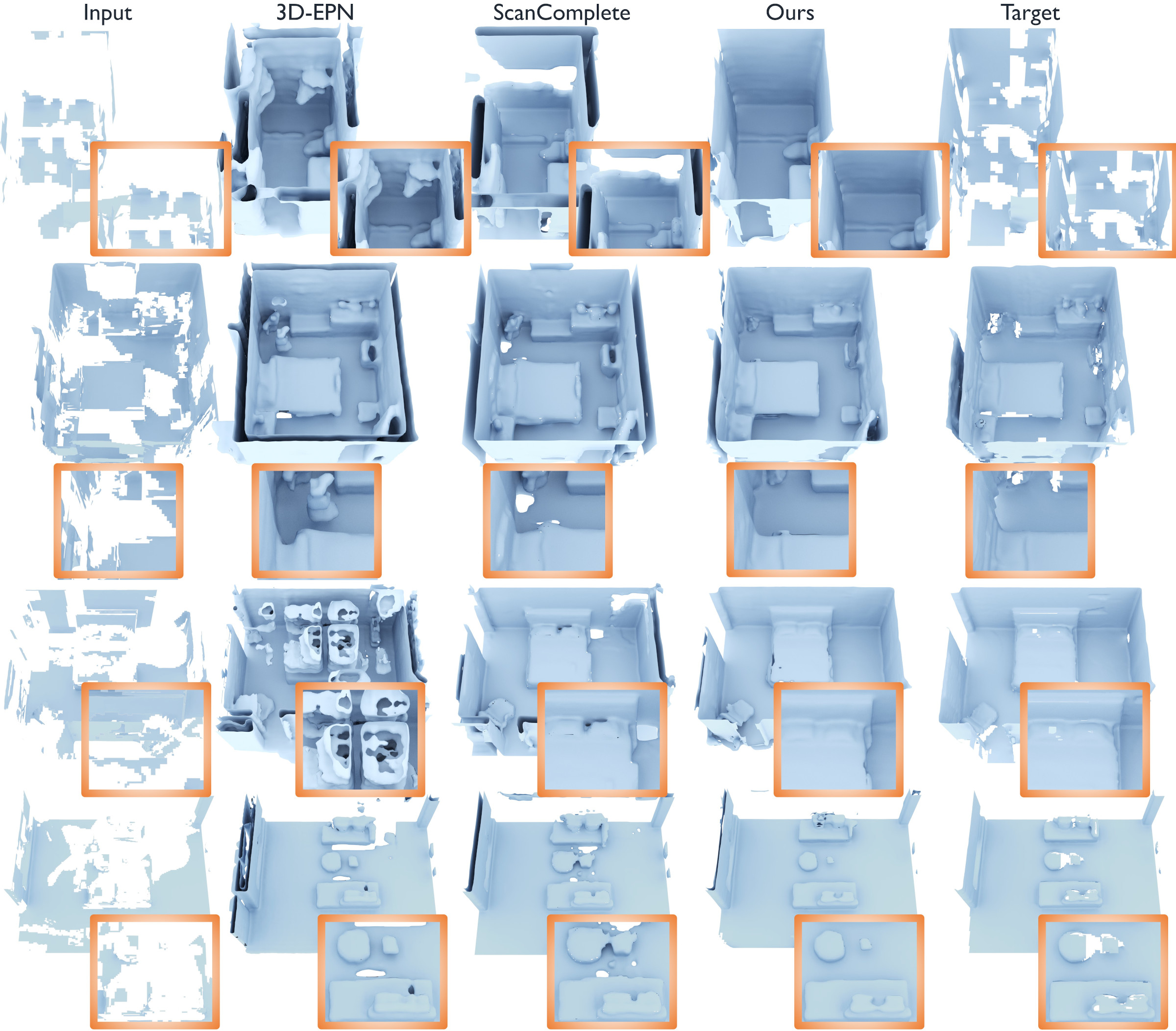}
	\vspace{-0.2cm}
	\caption{
	    Comparison to state-of-the-art scan completion approaches on Matterport3D~\cite{Matterport3D} data ($5$cm resolution), with input scans generated from a subset of frames. 
	    In contrast to the fully-supervised 3D-EPN~\cite{dai2017complete} and ScanComplete~\cite{dai2018scancomplete}, our self-supervised approach produces more accurate, complete scene geometry.
    }
	\label{fig:qualitative_real_5cm}
\end{figure*}

\begin{table*}[tp]
\begin{center}
	\small
	\begin{tabular}{| l || c | c | c | c | c |}
		\hline
		Method &  $\ell_1$ error & $\ell_1$ error & $\ell_1$ error & $\ell_1$ error \\ 
		 & entire volume & unobserved space & target & predicted  \\ \hline
        3D-EPN (unet)~\cite{dai2017complete} & 0.31 & 0.28 & 0.45 & 1.12 \\ \hline
        ScanComplete~\cite{dai2018scancomplete} & 0.20 & 0.15 & 0.51 & 0.74 \\ \hline
        Ours & {\bf 0.17} & {\bf 0.14} & {\bf 0.35} & {\bf 0.67} \\ \hline
	\end{tabular}
	\vspace{-0.1cm}
	\caption{Quantitative scan completion results on real-world scan data~\cite{Matterport3D}, with $\ell_1$ distance measured in voxel units for $5$cm voxels. Since target scans are incomplete, unobserved space in the target is masked out for all metrics.
	3D-EPN~\cite{dai2017complete} and ScanComplete~\cite{dai2018scancomplete} require full supervision, and so are trained on synthetic data~\cite{song2017ssc}.
	Despite their access to large quantities of synthetic 3D data, our self-supervised approach outperforms these methods while training solely on real-world data.
	}
	\label{tab:comparison_real}		
\end{center}
\vspace{-0.6cm}
\end{table*}

\section{Results and Evaluation}
\label{sec:results}

We evaluate our \MakeLowercase{\OURSFULL{}} on scene completion for RGB-D scans on both real-world scans where no fully complete ground truth is available~\cite{Matterport3D}, as well as in a supervised setting on synthetic scans which have complete ground truth information~\cite{song2017ssc}.
We use the train/test splits provided by both datasets: 72/18 and 5519/155 trainval/test scenes comprising 1788/394 and 39600/1000 rooms, respectively.
To measure completion quality, we follow \cite{dai2018scancomplete} and use an $\ell_1$ error metric between predicted and target TSDFs, where unobserved regions in the target are masked out.
Note that unsigned distances are used in the error computation to avoid sign ambiguities.
We measure the $\ell_1$ distance in voxel units of the entire volume (\emph{entire volume}), the unobserved region of the volume (\emph{unobserved space}), near the target surface (\emph{target}), and near the predicted surface (\emph{predicted}), using a threshold of $\leq 1$ to determine nearby regions, and a global truncation of $3$.
For all metrics, unobserved regions in the targets are ignored; note that on synthetic data  where complete ground truth is available, we do not have any unobserved regions to ignore.

\paragraph{Comparison to state of the art.}
In Table~\ref{tab:comparison_real}, we compare to several state-of-the-art approaches for scan completion on real-world scans from the Matterport3D dataset~\cite{Matterport3D}: the shape completion approach 3D-EPN~\cite{dai2017complete}, and the scene completion approach ScanComplete~\cite{dai2018scancomplete}.
These methods both require fully-complete ground truth data for supervision, which is not available for the real-world scenes, so we train them on synthetic scans~\cite{song2017ssc}.
Since 3D-EPN and ScanComplete use dense 3D convolutions, limiting voxel resolution, we use $5$cm resolution for training and evaluation of all methods.
Our self-supervised approach enables training on incomplete real-world scan data, avoiding domain transfer while outperforming previous approaches that leverage large amounts of synthetic 3D data.
Qualitative comparisons are shown in Figure~\ref{fig:qualitative_real_5cm}.

To evaluate our \OURS{} separate from its self-supervision, we also evaluate synthetic scan completion with full ground truth~\cite{song2017ssc}, in comparison to Poisson Surface Reconstruction~\cite{kazhdan2006poisson,kazhdan2013screened}, SSCNet\cite{song2017ssc}, 3D-EPN~\cite{dai2017complete}, and ScanComplete~\cite{dai2018scancomplete}.
All data-driven approaches are fully supervised, using input scans from \cite{dai2018scancomplete}.
Similar to the real scan scenario, we train and evaluate at $5$cm resolution due to resolution limitations of the prior learned approaches.
In Table~\ref{tab:comparison_synthetic}, we see that our sparse generative approach outperforms state of the art in a fully-supervised scenario.

\begin{table*}[tp]
\begin{center}
	\small
	\begin{tabular}{| l || c | c | c | c |}
		\hline
		Method &  $\ell_1$ error & $\ell_1$ error & $\ell_1$ error & $\ell_1$ error \\ 
		 & entire volume & unobserved space & target & predicted \\ \hline
        Poisson Surface Reconstruction~\cite{kazhdan2006poisson,kazhdan2013screened} & 0.53 & 0.51 & 1.70 & 1.18 \\ \hline
        SSCNet~\cite{song2017ssc} & 0.54 & 0.53 & 0.93 & 1.11 \\ \hline
        3D-EPN (unet)~\cite{dai2017complete} & 0.25 & 0.30 & 0.65 & 0.47  \\ \hline
        ScanComplete~\cite{dai2018scancomplete} & 0.18 & 0.23 & 0.53 & 0.42 \\ \hline
        Ours & {\bf 0.15} & {\bf 0.16} & {\bf 0.50} & {\bf 0.28}  \\ \hline
	\end{tabular}
	\vspace{-0.2cm}
	\caption{Quantitative scan completion results on synthetic scan data~\cite{song2017ssc}, where complete ground truth is available to supervise all data-driven approaches.
	$\ell_1$ distance is measured in voxel units for $5$cm voxels.
	\vspace{-0.4cm}
	}
	\label{tab:comparison_synthetic}		
\end{center}
\end{table*}

\begin{table*}[bp]
\begin{center}
	\small
	\begin{tabular}{| l || c | c | c | c | c |}
		\hline
		Method &  $\ell_1$ error & $\ell_1$ error & $\ell_1$ error & $\ell_1$ error \\ 
		 & entire volume & unobserved space & target & predicted\\ \hline
        Using crops for self-supervision & 0.13 & 0.09 & 1.25 & 0.68 \\ \hline
        Point cloud input & 0.15 & 0.09 & 1.82 & 0.92 \\ \hline
        Occupancy output & 0.13 & 0.10 & 0.89 & 0.86 \\ \hline
        2 hierarchy levels & 0.10 & 0.08 & 0.74 & 0.68 \\ \hline
        Ours & {\bf 0.09} & {\bf 0.07} & {\bf 0.71} & {\bf 0.60} \\ \hline
	\end{tabular}
	\vspace{-0.2cm}
	\caption{Ablation study of our self-supervision and generative model design choices on real-world scan data~\cite{Matterport3D}, with $\ell_1$ distance measured in voxel units for $2$cm voxels.
	}
	\label{tab:ablation_real}		
\end{center}
\end{table*}

\begin{figure*}[bp] 
	\centering
	\includegraphics[width=0.9\linewidth]{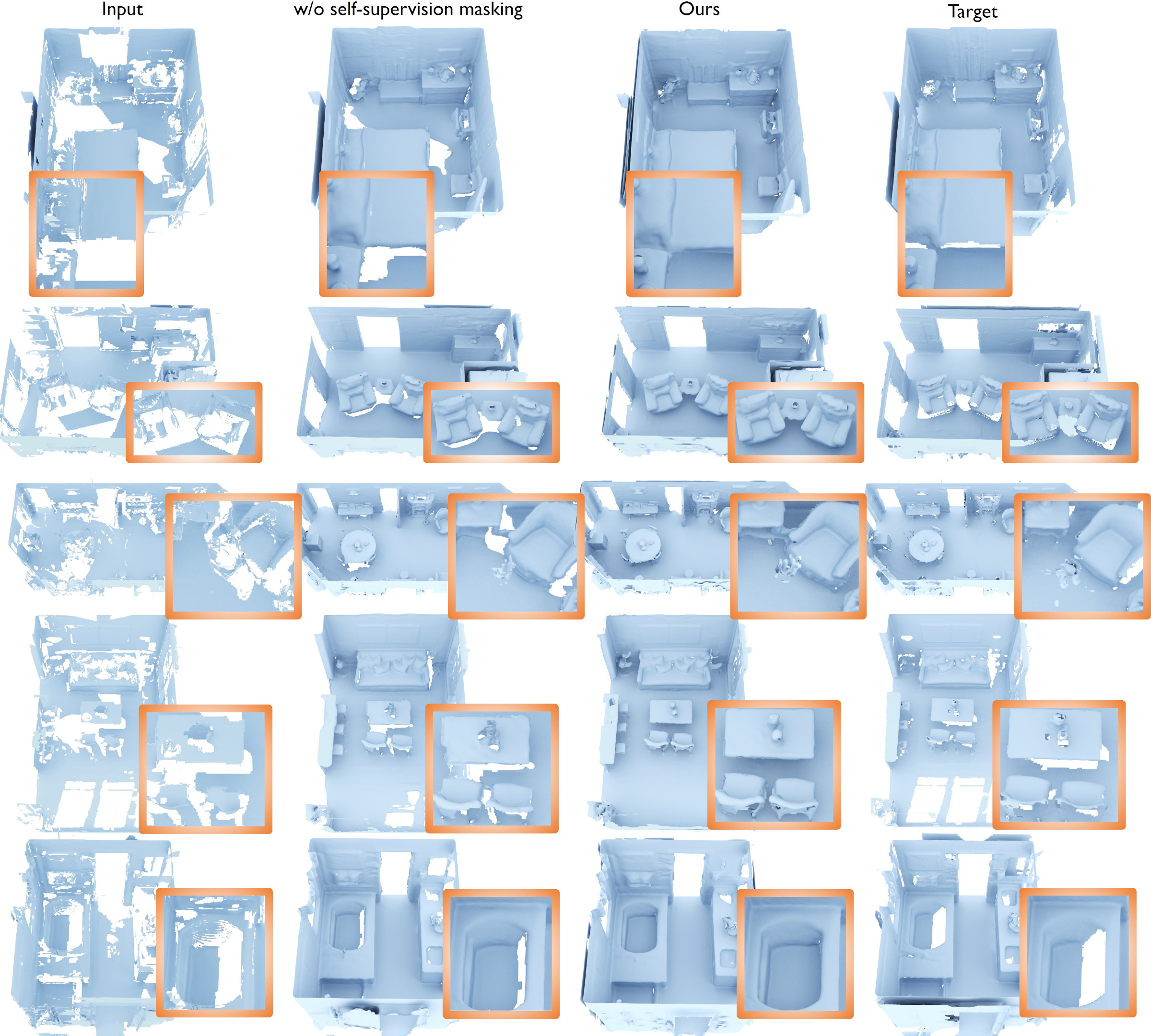}
	\caption{
	Scan completion results on Matterport3D~\cite{Matterport3D} data ($2$cm resolution), with input scans generated from a subset of frames. Our self-supervision approach using loss masking enables more complete scene prediction than direct supervision using the target RGB-D scan, particularly in regions where occlusions commonly occur.
    }
	\label{fig:qualitative_real_selfsupmasking}
\end{figure*}

\paragraph{Can self-supervision predict more complete geometry than seen during training?}
Our approach to self-supervision is designed to enable prediction of scene geometry beyond the completeness of the target scan data, by leveraging knowledge of observed and unobserved space in RGB-D scans.
To evaluate the completion quality of our method against the completeness of the target scene data, we perform a qualitative evaluation, as we lack fully complete ground truth to for quantitative evaluation.
In Figure~\ref{fig:qualitative_real_selfsupmasking}, we see that our completion quality can exceed the completeness of target scene data.
We additionally evaluate our approach with and without our self-supervision masking in Figure~\ref{fig:qualitative_real_selfsupmasking}, where \emph{w/o self-supervision masking} is trained using the same set of less-incomplete/more-incomplete scans but without the loss masking.
This can perform effective completion in regions commonly observed in target scans, but often fails to complete regions that are commonly occluded.
In contrast, our formulation for self-supervision using masking of unobserved regions enables predicting scene geometry even where the target scan remains incomplete.

\paragraph{Comparison of our self-supervision approach to masking out by random crops.}
In Table~\ref{tab:ablation_real}, we evaluate against another possible self-supervision approach: randomly cropping out target geometry to be used as incomplete inputs \emph{(using crops for self-supervision)}, similar to \cite{pathak2016context}.
This scenario does not reflect the data characteristics of real-world scan partialness (e.g., from occlusions and lack of visibility), resulting in poor completion performance.

\begin{figure}[!htb] 
	\centering
	\includegraphics[width=0.85\linewidth]{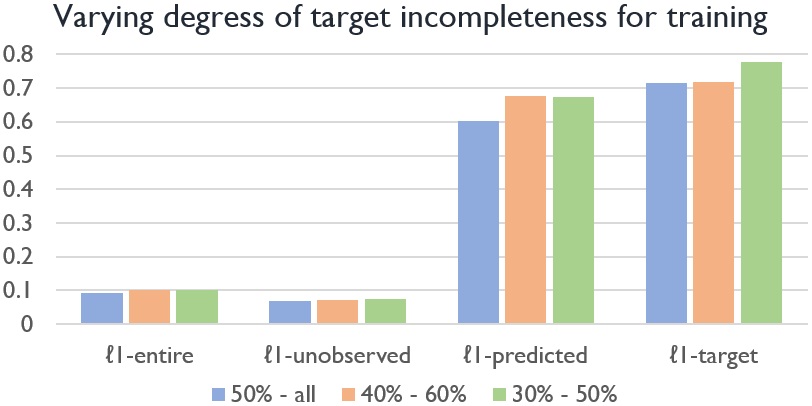}
	\vspace{-0.1cm}
	\caption{
	    Evaluating varying target data completeness available for training. We generate various incomplete versions of the Matterport3D~\cite{Matterport3D} scans using $\approx 30\%, 40\%, 50\%, 60\%$, and $100\%$ (all) of the frames associated with each room scene, and evaluate on the $50\%$ incomplete scans. 
	    Our self-supervised approach remains robust to the level of completeness of the target training data.
	\vspace{-0.3cm}
    }
	\label{fig:varying_incompleteness}
\end{figure}

\paragraph{What's the impact of the input/output representation?} 
In Table~\ref{tab:ablation_real}, we evaluate the effect of a point cloud input (vs. TSDF input), as well as occupancy output (vs. TSDF output).
We find that the TSDF representation has more potential descriptiveness in characterizing a surface (and its neighboring regions), resulting in improved performance in both input and output representation.

\paragraph{What's the impact of the degree of completeness of the target data during training?}
In Figure~\ref{fig:varying_incompleteness}, we evaluate the effect of the amount of completeness of the target data available for training.
We create several incomplete versions of the Matterport3D~\cite{Matterport3D} scans using varying amounts of the frames available: $\approx 30\%, 40\%, 50\%, 60\%$, and $100\%$ (all) of the frames associated with each room scene.
We train our approach using three different versions of input-target completeness: $50\%-all$ (our default), $40\%-60\%$, and $30\%-50\%$.
Even as the completeness of the target data decreases, our approach maintains robustness in predicting complete scene geometry.

\paragraph{Limitations}

Our \OURS{} approach for self-supervised scan completion enables high-resolution geometric prediction of complete geometry from real-world scans. 
However, to generate the full appearance of a 3D scene, generation and inpainting of color is also required.
Currently, our method also does not consider or predict the semantic object decomposition of a scene; however, we believe this would be an interesting direction, specifically in the context for enabling interaction with a 3D environment (e.g., interior redesign or robotic understanding).

%% file: 7conclusion.tex
\section{Conclusion}
\label{sec:conclusion}

In this paper, we presented a self-supervised approach for completion of RGB-D scan geometry that enables training solely on incomplete, real-world scans while learning a generative geometric completion process capable of predicting 3D scene geometry more complete than any single target scene seen during training.
Our sparse generative approach to generating a sparse TSDF representation of a surface enables much higher output geometric resolution than previous on large-scale 3D scenes.
Self-supervision allowing training only on real-world scan data for scan completion opens up new possibilities for various generative 3D modeling based only on real-world observations, perhaps mitigating the need for extensive synthetic data generation or domain transfer, and we believe this is a promising avenue for future research.

%% file: appendix.tex
\def\code#1{\texttt{#1}}

\section{\OURS{} Architecture Details}

\begin{figure*}[bp] 
	\centering
	\includegraphics[width=0.95\linewidth]{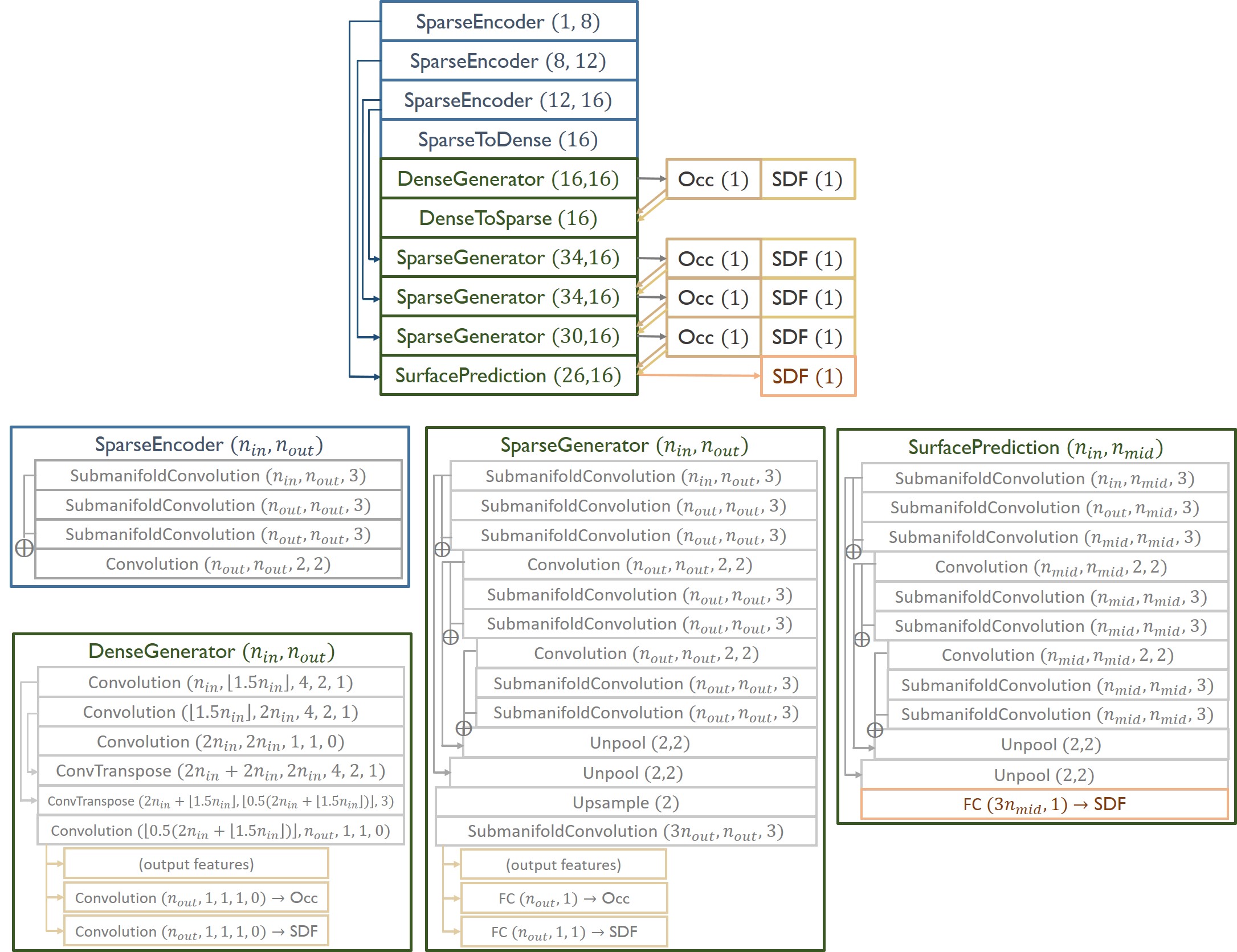}
	\vspace{-0.1cm}
	\caption{
	\OURS{} architecture in detail.
	The final TSDF values are highlighted in orange, and intermediate outputs in yellow.
	Convolution parameters are given as (nf\_in, nf\_out, kernel\_size, stride, padding), with stride and padding default to 1 and 0.
	Arrows denote concatenation, and $\bigoplus$ denotes addition.
    }
	\label{fig:architecture_details}
\end{figure*}

Figure~\ref{fig:architecture_details} details our \OURSFULL{} specification for scan completion.
Convolution parameters are given as (nf\_in, nf\_out, kernel\_size, stride, padding), with stride and padding default to 1 and 0 respectively.
Arrows indicate concatenation, and $\bigoplus$ indicates addition.
Each convolution (except the last) is followed by batch normalization and a ReLU.

\section{Varying Target Data Incompleteness}

\begin{figure*}[tp] 
	\centering
	\includegraphics[width=0.95\linewidth]{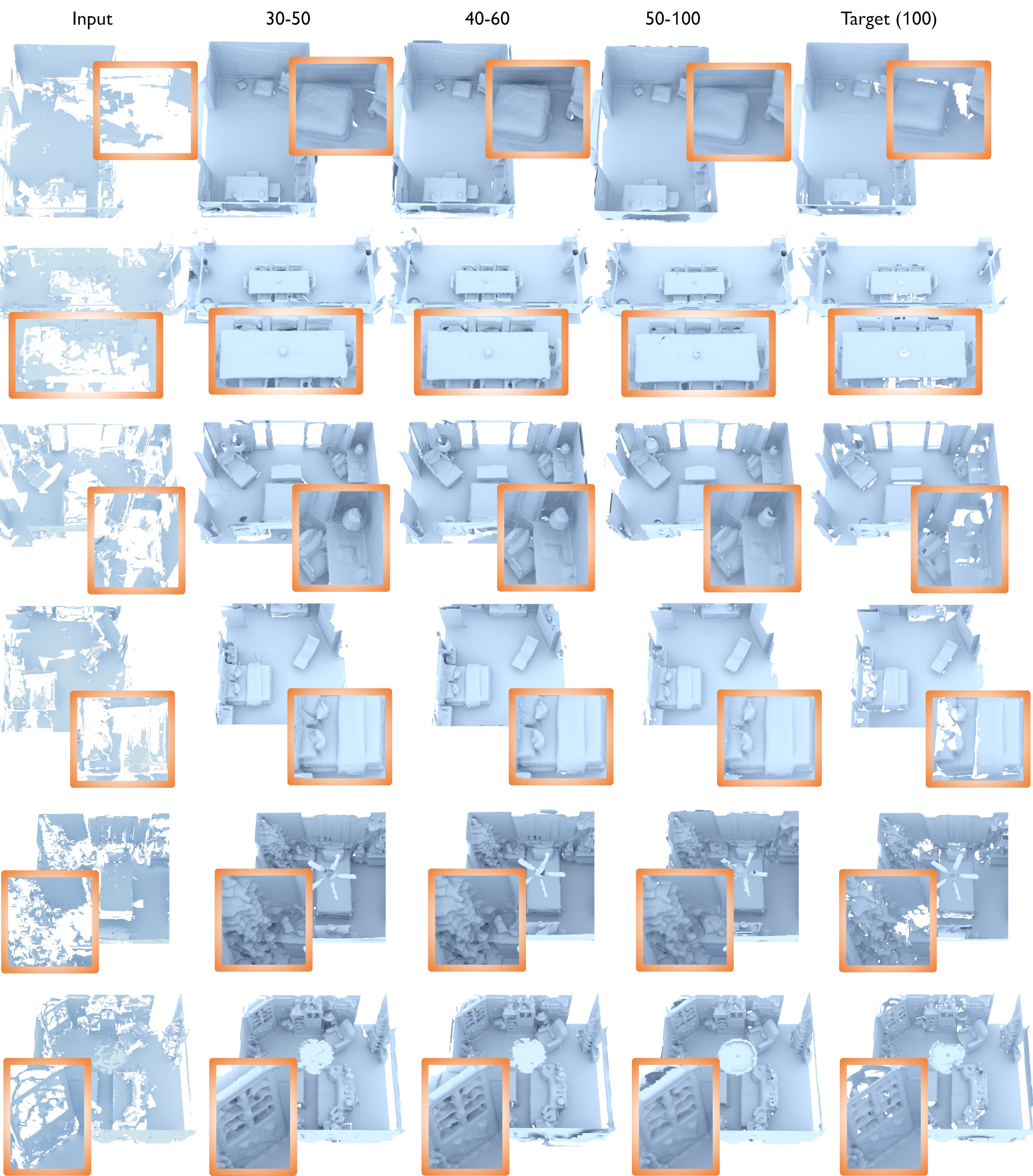}
	\caption{
	    Qualitative evaluation of varying target data completeness available for training. 
	    We generate various incomplete versions of the Matterport3D~\cite{Matterport3D} scans using $\approx 30\%, 40\%, 50\%, 60\%$, and $100\%$ of the frames associated with each room scene, and evaluate on the $50\%$ incomplete scans. 
	    Even as the level of completeness of the target data used during training decreases significantly, our self-supervised approach effectively learns the geometric completion process, maintaining robustness in generating complete geometry.
    }
	\label{fig:varying_incompleteness_visual}
\end{figure*}

Here, we aim to evaluate how well our self-supervision approach performs as the completeness of the target data seen during training decreases.
As long as there is enough variety in the completion patterns seen during training, our approach can learn to generate scene geometry with high levels of completeness.
To evaluate this, we generate several versions of target scans from the Matterport3D~\cite{Matterport3D} room scenes with varying degrees of completeness; that is, we use $\approx 50\%, 60\%$, and $100\%$  of the frames associated with each room scene  to generate three different levels of completeness in the target scans, using $\approx 30\%, 40\%$, and $50\%$ for the respective input scans.
We provide a quantitative evaluation in the main paper, and a  qualitative evaluation in Figure~\ref{fig:varying_incompleteness_visual}.
Even as the level of completeness in the target data used decreases, our approach maintains robustness its completion, informed by the deltas in incompleteness as to the patterns of generating complete geometry.

%% file: main.bbl
\begin{thebibliography}{10}\itemsep=-1pt

\bibitem{avetisyan2019scan2cad}
Armen Avetisyan, Manuel Dahnert, Angela Dai, Manolis Savva, Angel~X. Chang, and
  Matthias Nie{\ss}ner.
\newblock Scan2cad: Learning {CAD} model alignment in {RGB-D} scans.
\newblock In {\em {IEEE} Conference on Computer Vision and Pattern Recognition,
  {CVPR} 2019, Long Beach, CA, USA, June 16-20, 2019}, pages 2614--2623, 2019.

\bibitem{avetisyan2019end2end}
Armen Avetisyan, Angela Dai, and Matthias Niessner.
\newblock End-to-end cad model retrieval and 9dof alignment in 3d scans.
\newblock In {\em The IEEE International Conference on Computer Vision (ICCV)},
  October 2019.

\bibitem{Matterport3D}
Angel~X. Chang, Angela Dai, Thomas~A. Funkhouser, Maciej Halber, Matthias
  Nie{\ss}ner, Manolis Savva, Shuran Song, Andy Zeng, and Yinda Zhang.
\newblock Matterport3d: Learning from {RGB-D} data in indoor environments.
\newblock In {\em 2017 International Conference on 3D Vision, 3DV 2017,
  Qingdao, China, October 10-12, 2017}, pages 667--676, 2017.

\bibitem{shapenet2015}
Angel~X Chang, Thomas Funkhouser, Leonidas Guibas, Pat Hanrahan, Qixing Huang,
  Zimo Li, Silvio Savarese, Manolis Savva, Shuran Song, Hao Su, et~al.
\newblock Shapenet: An information-rich 3d model repository.
\newblock {\em arXiv preprint arXiv:1512.03012}, 2015.

\bibitem{choi2015robust}
Sungjoon Choi, Qian-Yi Zhou, and Vladlen Koltun.
\newblock Robust reconstruction of indoor scenes.
\newblock In {\em 2015 IEEE Conference on Computer Vision and Pattern
  Recognition (CVPR)}, pages 5556--5565. IEEE, 2015.

\bibitem{choy20194d}
Christopher~B. Choy, JunYoung Gwak, and Silvio Savarese.
\newblock 4d spatio-temporal convnets: Minkowski convolutional neural networks.
\newblock In {\em {IEEE} Conference on Computer Vision and Pattern Recognition,
  {CVPR} 2019, Long Beach, CA, USA, June 16-20, 2019}, pages 3075--3084, 2019.

\bibitem{curless1996volumetric}
Brian Curless and Marc Levoy.
\newblock A volumetric method for building complex models from range images.
\newblock In {\em Proceedings of the 23rd annual conference on Computer
  graphics and interactive techniques}, pages 303--312. ACM, 1996.

\bibitem{dahnert2019embedding}
Manuel Dahnert, Angela Dai, Leonidas~J. Guibas, and Matthias Niessner.
\newblock Joint embedding of 3d scan and cad objects.
\newblock In {\em The IEEE International Conference on Computer Vision (ICCV)},
  October 2019.

\bibitem{dai2017scannet}
Angela Dai, Angel~X. Chang, Manolis Savva, Maciej Halber, Thomas Funkhouser,
  and Matthias Niessner.
\newblock Scannet: Richly-annotated 3d reconstructions of indoor scenes.
\newblock In {\em The IEEE Conference on Computer Vision and Pattern
  Recognition (CVPR)}, July 2017.

\bibitem{dai2017bundlefusion}
Angela Dai, Matthias Nie{\ss}ner, Michael Zollh{\"{o}}fer, Shahram Izadi, and
  Christian Theobalt.
\newblock Bundlefusion: Real-time globally consistent 3d reconstruction using
  on-the-fly surface reintegration.
\newblock {\em {ACM} Trans. Graph.}, 36(3):24:1--24:18, 2017.

\bibitem{dai2017complete}
Angela Dai, Charles~Ruizhongtai Qi, and Matthias Nie{\ss}ner.
\newblock Shape completion using 3d-encoder-predictor cnns and shape synthesis.
\newblock In {\em 2017 {IEEE} Conference on Computer Vision and Pattern
  Recognition, {CVPR} 2017, Honolulu, HI, USA, July 21-26, 2017}, pages
  6545--6554, 2017.

\bibitem{dai2018scancomplete}
Angela Dai, Daniel Ritchie, Martin Bokeloh, Scott Reed, J{\"{u}}rgen Sturm, and
  Matthias Nie{\ss}ner.
\newblock Scancomplete: Large-scale scene completion and semantic segmentation
  for 3d scans.
\newblock In {\em 2018 {IEEE} Conference on Computer Vision and Pattern
  Recognition, {CVPR} 2018, Salt Lake City, UT, USA, June 18-22, 2018}, pages
  4578--4587, 2018.

\bibitem{graham20183dsemantic}
Benjamin Graham, Martin Engelcke, and Laurens van~der Maaten.
\newblock 3d semantic segmentation with submanifold sparse convolutional
  networks.
\newblock In {\em 2018 {IEEE} Conference on Computer Vision and Pattern
  Recognition, {CVPR} 2018, Salt Lake City, UT, USA, June 18-22, 2018}, pages
  9224--9232, 2018.

\bibitem{graham2017submanifold}
Benjamin Graham and Laurens van~der Maaten.
\newblock Submanifold sparse convolutional networks.
\newblock {\em arXiv preprint arXiv:1706.01307}, 2017.

\bibitem{han2017complete}
Xiaoguang Han, Zhen Li, Haibin Huang, Evangelos Kalogerakis, and Yizhou Yu.
\newblock High-resolution shape completion using deep neural networks for
  global structure and local geometry inference.
\newblock In {\em {IEEE} International Conference on Computer Vision, {ICCV}
  2017, Venice, Italy, October 22-29, 2017}, pages 85--93, 2017.

\bibitem{izadi2011kinectfusion}
Shahram Izadi, David Kim, Otmar Hilliges, David Molyneaux, Richard~A. Newcombe,
  Pushmeet Kohli, Jamie Shotton, Steve Hodges, Dustin Freeman, Andrew~J.
  Davison, and Andrew~W. Fitzgibbon.
\newblock Kinectfusion: real-time 3d reconstruction and interaction using a
  moving depth camera.
\newblock In {\em Proceedings of the 24th Annual {ACM} Symposium on User
  Interface Software and Technology, Santa Barbara, CA, USA, October 16-19,
  2011}, pages 559--568, 2011.

\bibitem{kazhdan2006poisson}
Michael~M. Kazhdan, Matthew Bolitho, and Hugues Hoppe.
\newblock Poisson surface reconstruction.
\newblock In {\em Proceedings of the Fourth Eurographics Symposium on Geometry
  Processing, Cagliari, Sardinia, Italy, June 26-28, 2006}, pages 61--70, 2006.

\bibitem{kazhdan2013screened}
Michael~M. Kazhdan and Hugues Hoppe.
\newblock Screened poisson surface reconstruction.
\newblock {\em {ACM} Trans. Graph.}, 32(3):29:1--29:13, 2013.

\bibitem{lorensen1987marching}
William~E. Lorensen and Harvey~E. Cline.
\newblock Marching cubes: {A} high resolution 3d surface construction
  algorithm.
\newblock In {\em Proceedings of the 14th Annual Conference on Computer
  Graphics and Interactive Techniques, {SIGGRAPH} 1987, Anaheim, California,
  USA, July 27-31, 1987}, pages 163--169, 1987.

\bibitem{maturana2015voxnet}
Daniel Maturana and Sebastian Scherer.
\newblock Voxnet: {A} 3d convolutional neural network for real-time object
  recognition.
\newblock In {\em 2015 {IEEE/RSJ} International Conference on Intelligent
  Robots and Systems, {IROS} 2015, Hamburg, Germany, September 28 - October 2,
  2015}, pages 922--928, 2015.

\bibitem{OccupancyNetworks}
Lars Mescheder, Michael Oechsle, Michael Niemeyer, Sebastian Nowozin, and
  Andreas Geiger.
\newblock Occupancy networks: Learning 3d reconstruction in function space.
\newblock In {\em Proceedings IEEE Conf. on Computer Vision and Pattern
  Recognition (CVPR)}, 2019.

\bibitem{newcombe2011kinectfusion}
Richard~A. Newcombe, Shahram Izadi, Otmar Hilliges, David Molyneaux, David Kim,
  Andrew~J. Davison, Pushmeet Kohli, Jamie Shotton, Steve Hodges, and Andrew~W.
  Fitzgibbon.
\newblock Kinectfusion: Real-time dense surface mapping and tracking.
\newblock In {\em 10th {IEEE} International Symposium on Mixed and Augmented
  Reality, {ISMAR} 2011, Basel, Switzerland, October 26-29, 2011}, pages
  127--136, 2011.

\bibitem{niessner2013hashing}
M. Nie{\ss}ner, M. Zollh\"ofer, S. Izadi, and M. Stamminger.
\newblock Real-time 3d reconstruction at scale using voxel hashing.
\newblock {\em ACM Transactions on Graphics (TOG)}, 2013.

\bibitem{Park2019DeepSDFLC}
Jeong~Joon Park, Peter Florence, Julian Straub, Richard~A. Newcombe, and Steven
  Lovegrove.
\newblock Deepsdf: Learning continuous signed distance functions for shape
  representation.
\newblock In {\em {IEEE} Conference on Computer Vision and Pattern Recognition,
  {CVPR} 2019, Long Beach, CA, USA, June 16-20, 2019}, pages 165--174, 2019.

\bibitem{pathak2016context}
Deepak Pathak, Philipp Kr{\"{a}}henb{\"{u}}hl, Jeff Donahue, Trevor Darrell,
  and Alexei~A. Efros.
\newblock Context encoders: Feature learning by inpainting.
\newblock In {\em 2016 {IEEE} Conference on Computer Vision and Pattern
  Recognition, {CVPR} 2016, Las Vegas, NV, USA, June 27-30, 2016}, pages
  2536--2544. {IEEE} Computer Society, 2016.

\bibitem{qi2017pointnet}
Charles~Ruizhongtai Qi, Hao Su, Kaichun Mo, and Leonidas~J. Guibas.
\newblock Pointnet: Deep learning on point sets for 3d classification and
  segmentation.
\newblock In {\em 2017 {IEEE} Conference on Computer Vision and Pattern
  Recognition, {CVPR} 2017, Honolulu, HI, USA, July 21-26, 2017}, pages 77--85,
  2017.

\bibitem{qi2017pointnetplusplus}
Charles~Ruizhongtai Qi, Li Yi, Hao Su, and Leonidas~J. Guibas.
\newblock Pointnet++: Deep hierarchical feature learning on point sets in a
  metric space.
\newblock In {\em Advances in Neural Information Processing Systems 30: Annual
  Conference on Neural Information Processing Systems 2017, 4-9 December 2017,
  Long Beach, CA, {USA}}, pages 5099--5108, 2017.

\bibitem{riegler2017octnetfusion}
Gernot Riegler, Ali~Osman Ulusoy, Horst Bischof, and Andreas Geiger.
\newblock Octnetfusion: Learning depth fusion from data.
\newblock In {\em 2017 International Conference on 3D Vision, 3DV 2017,
  Qingdao, China, October 10-12, 2017}, pages 57--66, 2017.

\bibitem{riegler2017OctNet}
Gernot Riegler, Ali~Osman Ulusoy, and Andreas Geiger.
\newblock Octnet: Learning deep 3d representations at high resolutions.
\newblock In {\em 2017 {IEEE} Conference on Computer Vision and Pattern
  Recognition, {CVPR} 2017, Honolulu, HI, USA, July 21-26, 2017}, pages
  6620--6629, 2017.

\bibitem{ronneberger2015u}
Olaf Ronneberger, Philipp Fischer, and Thomas Brox.
\newblock U-net: Convolutional networks for biomedical image segmentation.
\newblock In {\em International Conference on Medical image computing and
  computer-assisted intervention}, pages 234--241. Springer, 2015.

\bibitem{song2017ssc}
Shuran Song, Fisher Yu, Andy Zeng, Angel~X. Chang, Manolis Savva, and Thomas~A.
  Funkhouser.
\newblock Semantic scene completion from a single depth image.
\newblock In {\em 2017 {IEEE} Conference on Computer Vision and Pattern
  Recognition, {CVPR} 2017, Honolulu, HI, USA, July 21-26, 2017}, pages
  190--198, 2017.

\bibitem{ogn2017}
Maxim Tatarchenko, Alexey Dosovitskiy, and Thomas Brox.
\newblock Octree generating networks: Efficient convolutional architectures for
  high-resolution 3d outputs.
\newblock In {\em {IEEE} International Conference on Computer Vision, {ICCV}
  2017, Venice, Italy, October 22-29, 2017}, pages 2107--2115, 2017.

\bibitem{wang2017cnn}
Peng{-}Shuai Wang, Yang Liu, Yu{-}Xiao Guo, Chun{-}Yu Sun, and Xin Tong.
\newblock {O-CNN:} octree-based convolutional neural networks for 3d shape
  analysis.
\newblock {\em {ACM} Trans. Graph.}, 36(4):72:1--72:11, 2017.

\bibitem{wang2018adaptive}
Peng{-}Shuai Wang, Chun{-}Yu Sun, Yang Liu, and Xin Tong.
\newblock Adaptive {O-CNN:} a patch-based deep representation of 3d shapes.
\newblock {\em {ACM} Trans. Graph.}, 37(6):217:1--217:11, 2018.

\bibitem{Wang2017ShapeIU}
Weiyue Wang, Qiangui Huang, Suya You, Chao Yang, and Ulrich Neumann.
\newblock Shape inpainting using 3d generative adversarial network and
  recurrent convolutional networks.
\newblock In {\em {IEEE} International Conference on Computer Vision, {ICCV}
  2017, Venice, Italy, October 22-29, 2017}, pages 2317--2325, 2017.

\bibitem{whelan2015elasticfusion}
Thomas Whelan, Stefan Leutenegger, Renato~F. Salas{-}Moreno, Ben Glocker, and
  Andrew~J. Davison.
\newblock Elasticfusion: Dense {SLAM} without {A} pose graph.
\newblock In {\em Robotics: Science and Systems XI, Sapienza University of
  Rome, Rome, Italy, July 13-17, 2015}, 2015.

\bibitem{wu20153d}
Zhirong Wu, Shuran Song, Aditya Khosla, Fisher Yu, Linguang Zhang, Xiaoou Tang,
  and Jianxiong Xiao.
\newblock 3d shapenets: {A} deep representation for volumetric shapes.
\newblock In {\em {IEEE} Conference on Computer Vision and Pattern Recognition,
  {CVPR} 2015, Boston, MA, USA, June 7-12, 2015}, pages 1912--1920, 2015.

\end{thebibliography}
